\documentclass[10pt,twocolumn,letterpaper]{article}

\usepackage{cvpr}      % To produce the REVIEW version
\definecolor{cvprblue}{rgb}{0.21,0.49,0.74}
\usepackage[pagebackref,breaklinks,colorlinks,allcolors=cvprblue]{hyperref}

\def\paperID{1135} % *** Enter the Paper ID here
\def\confName{CVPR}
\def\confYear{2026}

\title{MPerS: Dynamic MLLM MixExperts Perception-Guided Remote Sensing Scene Segmentation}

\author{
	Ziyi Wang$^{1}$, Xianping Ma$^{2}$, Ziyao Wang$^{1}$, Hongyang Zhang$^{1}$, Man On Pun$^{1\dagger}$\\
	$^{1}$The Chinese University of Hong Kong (Shenzhen) \quad $^{2}$Southwest Jiaotong University\\
	{\tt\footnotesize \{ziyiwang8, ziyaowang2\}@link.cuhk.edu.cn, 
	mxp@swjtu.edu.cn, simonpun@cuhk.edu.cn}
}

\usepackage{multirow}
\usepackage{cuted}  % 
\usepackage{capt-of} % 

\usepackage[table]{xcolor}
\begin{document}

\maketitle

\begin{strip}
	\centering
	\includegraphics[width=1\textwidth]{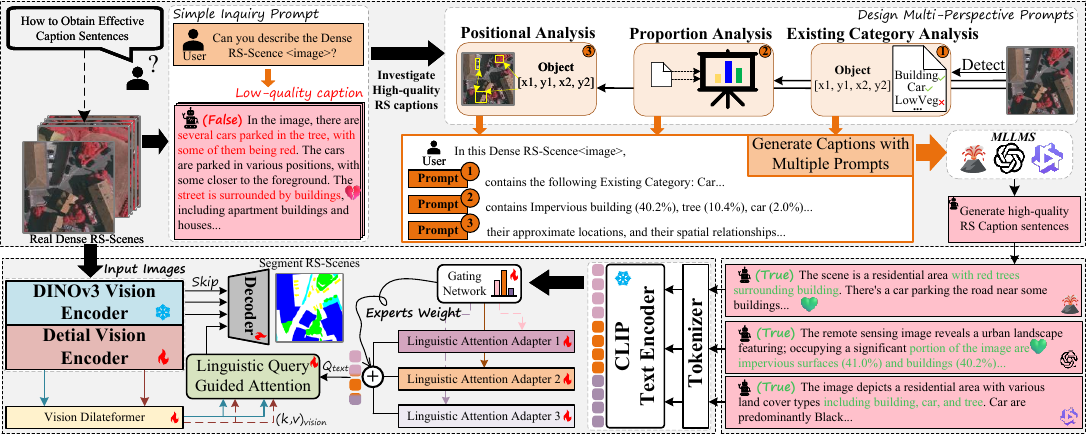}
	\captionof{figure}{Workflow of MPerS. 
		Simple prompts may generate inappropriate captions, leading to erroneous perceptual understanding. This motivates studying effective MLLM caption perception strategies.
		First, the characteristics of the RS-Scene are analyzed, and three sets of prompts are generated accordingly. These prompts are then fed into Multimodal Large Language Models (MLLMs) to perceive the scene and generate corresponding captions. The captions are then encoded via the CLIP text encoder to construct a dynamic gating network. Finally, through guided attention, textual information modulates visual features, which are then decoded into the segmentation map.}
	\label{fig:flow}
\end{strip}

\begin{abstract}
The multimodal fusion of images and scene captions has been extensively explored and applied in various fields.
However, when dealing with complex remote sensing (RS) scenes, existing studies have predominantly concentrated on architectural optimizations for integrating textual semantic information with visual features, while largely neglecting the generation of high-quality RS captions and the investigation of their effectiveness in multimodal semantic fusion. 
In this context, we propose the Dynamic MLLM Mixture-of-Experts Perception-Guided Remote Sensing Scene Segmentation, referred to as MPerS.
We design multiple prompts for MLLMs to generate high-quality RS captions, enabling MLLMs to perceive RS scenes from diverse expert perspectives. 
DINOv3 is employed to extract dense visual representations of land-covers.
We design a Dynamic MixExperts module that adaptively integrates the most effective textual semantics. Linguistic Query Guided Attention is constructed to utilize textual semantic information to guide visual features for precise segmentation. The MLLMs include LLaVA, ChatGPT, and Qwen. Our method achieves superior performance on three public semantic segmentation RS datasets.
\end{abstract}
\section{Introduction}
Semantic segmentation \cite{kirillov2023segment} is a fundamental task in computer vision. In remote sensing scenes (RS-Scence) with high resolution and extensive spatial coverage, it serves as a crucial technique supporting various remote sensing (RS) interpretation applications. Semantic segmentation in RS scenes assigns each pixel to a specific semantic category.
In traditional research, semantic segmentation in RS-Scence has mainly relied on single-modality high-resolution imagery \cite{long2015fully,ronneberger2015u}. As a result, most existing studies have concentrated on effectively extracting land-cover features to facilitate segmentation in downstream tasks.
However, due to the complexity of RS-Scenes and the dense distribution of land-covers, single-modality image data requires a large number of precise manual annotations \cite{waqas2019isaid}, which are time-consuming and costly.

Furthermore, single-modality image data provide a very limited perception of RS-Scence, thereby imposing substantial constraints on the performance of models trained exclusively on such data. From the perspective of human scene perception, we likewise integrate multiple heterogeneous sensory modalities to comprehend and analyze the inherent characteristics and constituent elements of the current scene. Consequently, accurate perception and fusion of multimodal data better represent real-world scenes and are highly applicable to practical RS interpretation.

The diversity of sensors and the expanding volume of RS imagery have also increasingly driven interest in multimodal approaches for RS semantic segmentation. Perception in RS-Scence has expanded from single-image data sources to multispectral imagery \cite{kemker2018algorithms}, synthetic aperture radar data \cite{li2024sardet}, and digital surface model \cite{vaihingen20182d} , enabling scene understanding from multiple perspectives. Researchers have developed various fusion strategies to leverage information from these data sources.

In recent years, visual-language representation learning \cite{li2025dynamic,zhu2025skysense} has emerged as an important research area in computer vision. Its goal is to leverage deep learning techniques to extract unified cross-modal feature representations from image-text pairs, thereby enhancing various multimodal tasks. 
Advances in multimodal large language models (MLLMs) have further propelled the field, enabling their application to diverse downstream tasks \cite{zhu2025segagent}.
For dense RS-Scenes, the fusion of textual semantic features with visual features is undoubtedly important. The way textual captions are generated for perceiving scenes, as well as their effectiveness, directly impacts the performance of segmentation after visual-textual fusion.

To address the aforementioned issues, we propose MPerS, a DINOv3-based semantic segmentation method, where the segmentation is guided by captions generated by diverse MLLM experts, each of whom perceives remote sensing scene characteristics at varying levels of domain expertise. Previous methods \cite{wang2024metasegnet} rely solely on simple descriptive or category texts to perceive RS scenes. Our work aims to generate high-quality captions that provide textual semantic information conducive to downstream segmentation tasks and incorporates Guided Attention fusion between images and text features. The workflow of our work is show in Fig.~\ref{fig:flow}. The main contributions are as follows:
\begin{itemize}
	\item
	We propose MPerS. To our knowledge, this is the first work to investigate the effectiveness of text-guided captions generated by different MLLMs in aiding multimodal dense RS-Scene segmentation.
	\item
	Three types of MLLMs are leveraged to build a Dynamic MixExperts text encoder, designed to extract the textual semantic information that most effectively guides image features for downstream segmentation tasks.
	\item
	Linguistic Query Guided Attention is designed to fully utilize textual information, ensuring that textual semantics guide image features for more accurate segmentation. Our method demonstrates excellent performance on three public RS datasets: Potsdam, Vaihingen, and SynDrone.
\end{itemize}

\section{Related Work}

\subsection{Semantic Segmentation for Remote Sensing Scenes}
With the increasing application of deep learning-based semantic segmentation in remote sensing Sences (RS-Scence), many approaches have achieved remarkable performance.
For instance, multi-scale Convolutional Neural Networks (CNNs) have been designed to efficiently extract land-cover features \cite{li2020scattnet,wang2021scale}, while various attention mechanisms have been tailored to the unique characteristics of RS-Sences  \cite{li2021multistage,li2021abcnet}. Additionally, several studies have proposed hybrid architectures that combine Transformers and CNNs \cite{wang2022unetformer,song2022ctmfnet,he2022swin, gao2021stransfuse, zhang2022transformer,liu2024esms}, aiming to achieve fine-grained land-cover segmentation. The aforementioned methods have achieved excellent segmentation performance. However, obtaining pixel-level annotations for RSIs is time-consuming and labor-intensive, and the segmentation performance remains significantly constrained. Accordingly, multimodal methods \cite{ma2024sam,ma2024manet} have garnered increasing attention and enhance the perception of RS scenes.

\subsection{Remote Sensing Scenes Captioning}
Unlike natural scenes, RS-Scence contain a large number of densely and complexly distributed land-cover objects \cite{lei2021hybrid}. RS-Scene captions should not only convey the categories and quantities of land-cover objects, but also precisely capture the spatial relationships and interactions between different land-cover types. By appropriately designing text-assisted tasks, the visual features extracted by the model can be effectively optimized.
Zhao et al. \cite{zhao2021high} proposed a network capable of generating pixel-level segmentation results for both key objects and image descriptions. 
Ye et al. \cite{ye2022joint} introduced a multi-label classification task to incorporate prior knowledge, which is then fed into the decoder to improve the accuracy of generated captions.
Furthermore, Zhao and Xiong \cite{zhao2024cooperative} manually annotated object regions in the RS-Scenes Captioning dataset and refined the decoder to enable the model to more effectively utilize fine-grained and regional features.
Multimodal Large language model (MLLM) techniques \cite{ouyang2022training} offer significant advantages in the field of natural language processing.
Hu et al. \cite{hu2025rsgpt} proposed a large-scale remote sensing model for RS-Scenes Captioning and RS visual question answering tasks, and also provided a fine-grained RS-Scenes dataset. 
Yang et al. \cite{yang2024bootstrapping} proposed a two-stage vision-language training method to align visual features with textual information.
Compared to traditional caption setting methods, leveraging MLLMs can produce high-quality captions for RS-Scenes.

\subsection{Vision-Language Research in Remote Sensing}
In previous work, learning paradigms for computer vision tasks have predominantly followed pretraining and fine-tuning strategy. Models are initially pretrained on natural images and then fine-tuned on downstream RS tasks, demonstrating excellent performance \cite{guo2025dinomaly, jose2025dinov2, xu2022groupvit}. Building on this, an increasing number of researchers have combined multimodal learning with this approach for computer vision tasks\cite{zhang2024segclip}. 
Lu et al. \cite{lu2019vilbert} used a co-attention mechanism to capture correlations between modalities and learn joint visual-language representations.
Li et al. \cite{li2021align} proposed a learning method that combines visual and language representations, using momentum distillation to pre-align them and improve information fusion from the two modalities.
Huo et al. \cite{huo2021wenlan} proposed a large-scale multimodal pretraining method to effectively bridge vision and language via joint image-text training.
FILM \cite{zhao2024image} leverages textual semantics to guide multi-task image fusion, enhancing feature understanding and fusion effectiveness. 
Due to the heterogeneity of RS-Sences data, the application of Vision-Language model in RS remains largely underexplored. 
RemoteCLIP \cite{liu2024remoteclip} established a unified RS vision-language model and demonstrated its effectiveness on image classification and image-text retrieval tasks.
Rao et al. \cite{rao2022denseclip} Utilized the CLIP model to classify pixels, achieving denser and more precise semantic segmentation by transforming image-text matching into pixel-text matching.

\begin{figure}[!t]  % 使用 dblfloatfix
	\centering
	\includegraphics[width=0.5\textwidth]{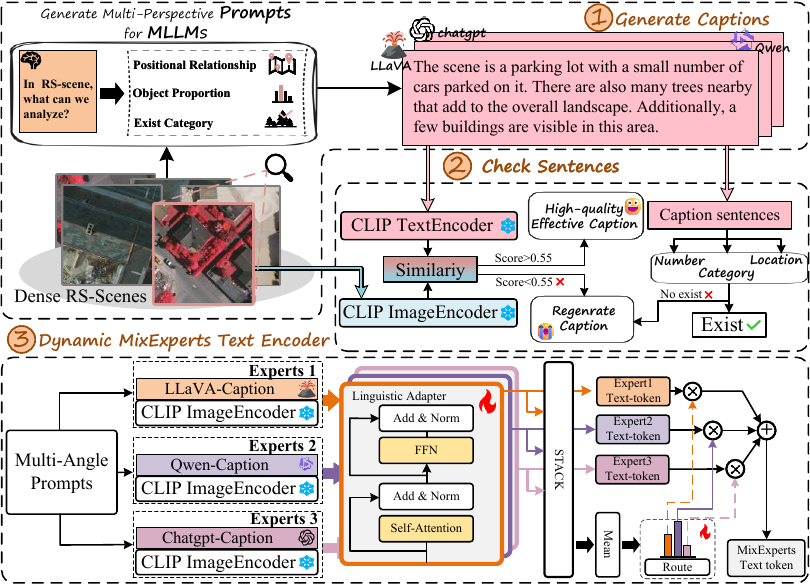}
	\caption{The pipeline for effective semantic text acquisition has three stages: 1) Obtain captions of the scene from MLLMs using multi-perspective prompts. 2) Establish a checking strategy to guarantee captions’ effectiveness. 3) Design a Dynamic MixExperts gating network to extract textual semantic information most beneficial for downstream segmentation tasks.}
	\label{fig:pipeline}
\end{figure}

\begin{figure*}[!t]  % 使用 dblfloatfix
	\centering
	\includegraphics[width=1\textwidth]{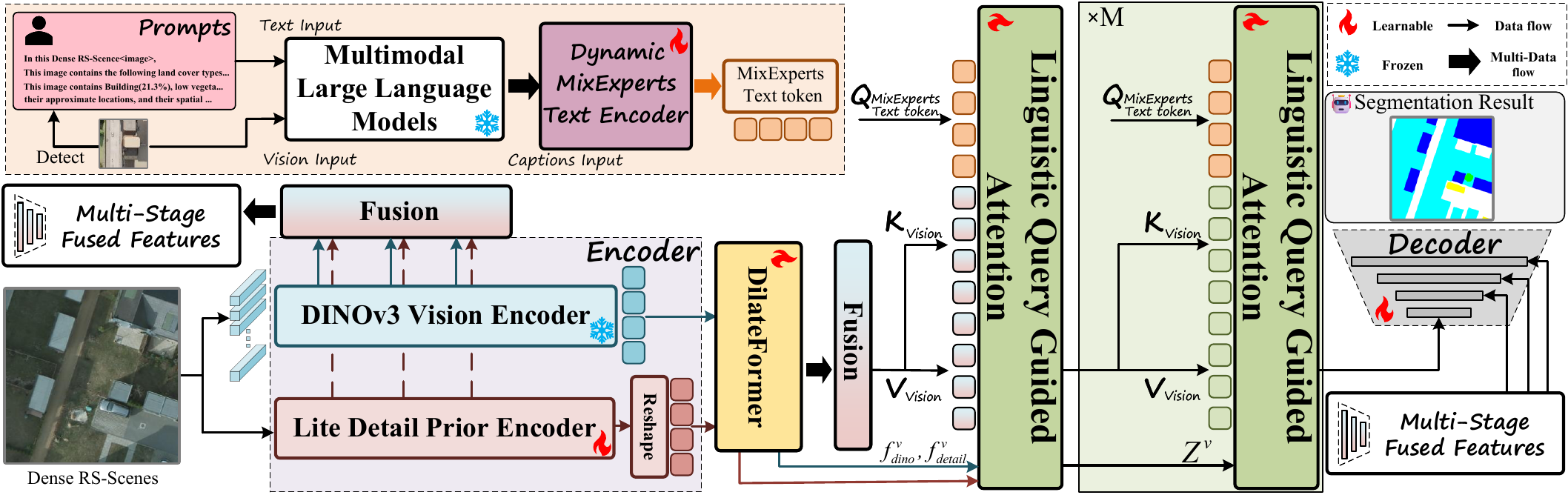}
	\caption{The framework of MPerS, which encompasses four units: vision encoder, Dynamic MLLM MixExperts extract effective textual semantic features, Linguistic Query Guided Attention to fuse textual and visual features, segmentation decoder.}
	\label{fig:network}
\end{figure*}

\section{Method}

Section \ref{sec:workflow} details the overall workflow of MPerS. Section \ref{sec:Captioning} presents the generation and verification strategy for high-quality RS-Scene captions produced by MLLMs, while Section \ref{sec:Framework} further presents the overall framework and component design of MPerS.

\subsection{Workflow Overview}
\label{sec:workflow}

Fig.~\ref{fig:flow} illustrates the detailed workflow of MPerS. The workflow of our work consists of two parts: (i) Generating high-quality RS-Scene captions using MLLMs. (ii) Design of the RS-Scene segmentation framework. 

Although single-modality RS image semantic segmentation methods have shown significant success, their performance is inherently limited. However, in multimodal methods\cite{wu2025fsvlm,hu2025ringmo} that leverage generated RS captions, the quality of the captions is critical.
To ensure that generated captions accurately describe RS-Scenes, we analyze the complexity of the Dense RS-Scenes and design multiple prompts as input to MLLMs for generating high-quality RS captions. 
Subsequently, As shown in Fig.~\ref{fig:pipeline}, MLLMs are employed to produce RS captions. These captions are encoded using CLIP Text encoder \cite{radford2021learning}, while a Dynamic MixExperts text encoder adjusts the contribution of each MLLM expert during training to extract the most effective textual semantic information for dense RS-Scene segmentation.
We use DINOv3 \cite{simeoni2025dinov3} and construct a Detailed Vision Encoder for visual feature encoding. A Linguistic Query Guided Attention module is designed to fuse textual and visual features, followed by multi-stage skip decoding to segment dense RS-Scenes.

\subsection{RS-Sences Captioning with MLLMs} 
\label{sec:Captioning}

High-quality RS-Scene captions should accurately reflect key land-cover categories and their spatial relationships. Traditional RS captions are constructed from class texts, ignoring spatial information. Captions generated by MLLMs using simple Prompt also may include hallucinated land-cover elements and lack fine-grained details.

To obtain high-quality RS-Scene captions, as shown in Fig.~\ref{fig:flow}, we desgin three types of prompts from different perspectives as input to MLLM, primarily including the analysis of existing land-covers, classes proportions, and location relationship. Meanwhile, Multiple MLLM experts simultaneously perceive the RS-Scene and generate multi-captions ${P_c}:[caption{\rm{ }}1,{\rm{ }}caption{\rm{ }}2,{\rm{ }}...]$.

To ensure the effectiveness of captions for downstream tasks, we design an RS-Scene Caption Sentences Check strategy. As shown in Fig.~\ref{fig:pipeline}, a similarity matrix is constructed between captions and the corresponding RS images, with a score threshold set to $\tau=0.55$. When the score below $\tau$, captions are regenerated, and the sentences are checked to ensure they contain the three key elements: Number, category, and location relationship.

\subsection{MPerS Framework}
\label{sec:Framework}

\subsubsection{Vision Feature Extraction}

For complex and densely segmented RS-Scene images $I \in \mathbb{R}^{H \times W \times 3}$, as shown in Fig.~\ref{fig:network}, we employ DINOv3 for visual feature encoding. Meanwhile, we design a \textbf{L}ite \textbf{D}etail \textbf{P}rior \textbf{E}ncoder with CNN blocks to incorporate RS-Scene detail priors and fine land-cover features.
\begin{align}
	&f_{{\rm{dino}}}^v = \mathrm{DINOv3Encoder}(I), \\
	&\hspace{0.2cm}f_{{\rm{detail}}}^v = \mathrm{LDPEncoder}(I). 
\end{align}

For a RS-Scene image $I$, deep visual features $\{f_{{\rm{dion}}}^v,f_{{\rm{detial}}}^v\}$ can be extracted. The visual features are then fed into a single-dilation DilateFormer \cite{jiao2023dilateformer}, followed by fusion to obtain fused features $f_{{\rm{F_d}}}^v$. Multi-level features are particularly important for RS tasks. As shown in Fig.~\ref{fig:network}, intermediate features from DINOv3 and the LDP Encoder are extracted and fused for subsequent segmentation decoding. After fusion, three sets of skip connection features  $f_{F_i}^v,{\rm{ }}i = 1,{\rm{ }}2,{\rm{ }}3$ are obtained.

\subsubsection{Dynamic MixExperts Text Encoder}
Inspired by Sa2VA \cite{yuan2025sa2va} and InstanceCap\cite{fan2025instancecap}, we investigate whether generating prompts from multiple perspectives of RS-Scenes and inputting them into MLLMs can produce higher quality captions. Our idea was validated by subsequent experiments. Given a RS-Scene $I \in \mathbb{R}^{H \times W \times 3}$, as shown in Fig.~\ref{fig:pipeline}, we employ multi-perspective prompts to input $I$ into three types of MLLMs, generating three sets of corresponding captions. Next, the three generated caption sets are input into a CLIP text encoder with frozen parameters to produce multiple sets of text tokens $\{ \Phi _{{\rm{MLLM}}1}^t,\Phi _{{\rm{MLLM}}2}^t,\Phi _{{\rm{MLLM}}3}^t\}$. 

To guarantee the effectiveness of semantic text information in guiding image segmentation, we design a Dynamic MixExperts Text Encoder that leverages multiple sets of text tokens generated by MLLMs. 
Using a dynamic routing mechanism, each expert exploits its domain-specific knowledge to interpret the current remote sensing scene and produces routing weights that best support the downstream segmentation task. The formulation is as follows:
\begin{align}
	& \hspace{0.4cm} G_m = \sigma \big( g(\frac{1}{M} \sum_{n=1}^{M}  \Phi _{{\rm{MLLM}}_n}^t \big)_m , (n = 1, 2, 3), \\
	& {T}_{\text{MixExperts}} = \sum_{m=1}^{M} W_m \, G_m \, \Phi _{{\rm{MLLM}}_m}^t, (m = 1, 2, 3).
\end{align}

Before being fed into the gating network, $\Phi _{{\rm{MLLM}}_n}^t$ is adapted by a Linguistic Attention. $g( \cdot )$ denotes the gating network. $\sigma ( \cdot )$ represents the sigmoid activation function. $G_m$ denotes the dynamic gating value corresponding to the m-th expert. $W_m$ denotes the expert weight, Here,$n$ and $m$ index the $M$ experts.
Finally, we obtain the MixExpert text token ${T}_{\text{MixExperts}}$.

\begin{figure}[!t]  % 使用 dblfloatfix
	\centering
	\includegraphics[width=0.45\textwidth]{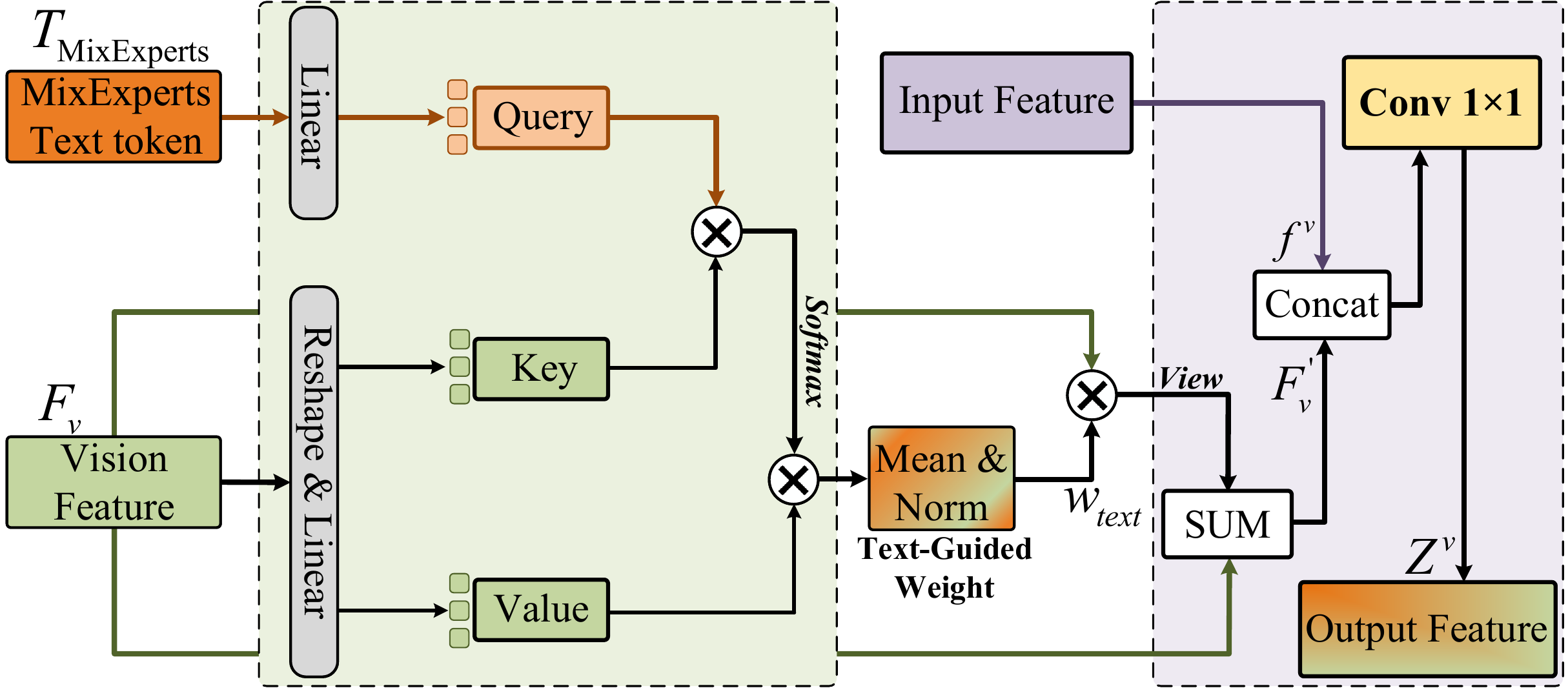}
	\caption{Architecture of the Linguistic Query Guided Attention.}
	\label{fig:QGCA}
\end{figure}

\subsubsection{Linguistic Query Guided Attention}
In the context of complex and densely structured RS-Scenes, it is crucial to enable models to achieve accurate multimodal perception and information fusion.
As shown in Fig.~\ref{fig:QGCA}, we design a \textbf{L}inguistic \textbf{Q}uery \textbf{G}uided \textbf{A}ttention that enables text features to provide more detailed guidance for visual perception and better retain the original visual feature information during the fusion process.

Initially, we map the extracted visual features $F_v$ and the text token ${T}_{\text{MixExperts}}$ into a shared embedding space. A Guided Attention is then applied, using text tokens as queries $Q_{text}$ and visual tokens as keys $K_{vision}$ and values $V_{vision}$, to produce the text-guided visual features $F'_v$  as:
\begin{align}
	& w_{text} = \xi (softmax(\frac{{{Q_{text}}{K_{vision}}^T}}{{\sqrt d }}) \cdot {V_{vision}}), \\
 	&\hspace{2cm} F'_v = \mathrm{view}(w_{text} \cdot F_v) + F_v,
\end{align}
where $\xi$ denotes the mean and normalization operation for generating the weights, and $w_{text}$ represents the text-guided weights. The generated visual feature $F'_v$ is then fused with the original vision feature $f^v$ as follows:
\begin{align}
	{Z^v} = {\rm{Con}}{{\rm{v}}_{1 \times 1}}(concat(F'_v,{\rm{ }}{f^v})),
\end{align}
where ${Z^v}$ denotes the output text-guided visual feature. As shown in Fig.~\ref{fig:network}, during the initial input to QGCA, ${f^v}$ refers to the deep visual features $\{f_{{\rm{dion}}}^v,f_{{\rm{detial}}}^v\}$ obtained from the encoded image. In subsequent modules, ${f^v}$ corresponds to the text-guided visual feature ${Z^v}$.

\subsubsection{Decoding Module}
Finally, the text-guided visual feature ${Z^v}$ and the multi-stage fused features $f_{F_i}^v,{\rm{ }}i = 1,{\rm{ }}2,{\rm{ }}3$ obtained are fed into the vision decoder. The decoder is constructed following a U-Net \cite{wang2022unetformer} architecture, producing the RS-Scene segmentation map, denoted as $F_{seg} = D(Z^v,f_{F_1},f_{F_2},f_{F_3})$, where $F_{seg}$ denotes the segment result. For the loss function, we use only the cross-entropy loss $\mathcal{L}_{ce}$ \cite{mao2023cross} for training, which yields satisfactory results. 

\section{Experiments}

\begin{table*}[t!]
	\centering
	\caption{Quantitative comparison with state-of-the-art methods on the Vaihigen dataset. $*$ denotes DINOv3 weights with 7B parameters; entries without $*$ use 0.3B distilled weights. FILM employs a self-designed backbone.}
	\resizebox{\linewidth}{!}{
		\begin{tabular}{c|c|ccccc|cc}
			\toprule
			\multirow{2}[4]{*}{Model} & \multirow{2}[4]{*}{Backbone} & \multicolumn{5}{c|}{Per-class IoU(\%)/F1(\%)} & \multirow{2}[4]{*}{mIoU(\%)} & \multirow{2}[4]{*}{mF1(\%)} \\
			\cmidrule{3-7}          &       & Impervious surface & Building & Low vegetation & Tree & Car  &       &  \\
			\midrule
			MAResUnet & ResNet-34 & 79.23/88.22 & 86.47/92.74 & 58.41/73.74 & 75.30/85.91 & 56.19/71.95 & 71.12 & 81.61 \\
			Unetformer & ResNet-18 & 76.19/86.49 & 84.63/91.67 & 55.63/71.49 & 74.42/85.33 & 51.68/68.14 & 68.51 & 80.63 \\
			DC-Swin & Swin-B & 72.78/84.24 & 81.72/89.94 & 52.86/69.16 & 73.05/84.42 & 48.69/65.49 & 65.82 & 77.35 \\
			A$^2$-FPN & ResNet-18 & 79.59/88.64 & 88.50/93.90 & 59.30/74.45 & 75.77/86.22 & 65.34/79.04 & 73.82 & 84.45 \\
			RS$^3$Mamba & VMamba-T & 79.87/88.81 & 88.76/94.05 & 60.35/75.27 & 76.56/86.72 & 64.85/78.68 & 74.08 & 84.71 \\
			FILM  & -     & 70.40/82.63 & 78.14/87.73 & 51.89/68.33 & 73.79/84.92 & 40.63/57.78 & 62.97 & 76.28 \\
			MetaSegNet & Swin-B & 73.41/84.67 & 82.03/90.13 & 55.96/71.76 & 73.03/84.41 & 47.49/64.39 & 66.38 & 79.07 \\
			SegCLIP & Swin-B & 76.88/86.93 & 85.17/91.99 & 59.20/74.37 & 76.19/86.49 & 63.71/77.84 & 72.23 & 81.27 \\
			\midrule
			MPerS (Ours)  & DINOv3-SAT & 79.58/88.63 & 88.22/93.74 & 61.54/76.19 & 78.05/87.67 & 65.27/78.99 & 74.53 & 85.04 \\
			MPerS (Ours) & DINOv3-LVD & \underline{81.02}/\underline{89.52} & \underline{89.55}/\underline{94.49} & \underline{63.76}/\underline{77.87} & \underline{79.08}/\underline{88.32} & \underline{72.09}/\underline{83.78} & \underline{77.10} & \underline{86.79} \\
			\midrule
			MPerS$^*$ (Ours) & DINOv3-SAT & 79.40/88.52 & 88.75/94.04 & 61.56/78.60 & 78.60/88.02 & 69.74/82.17 & 75.61 & 85.79 \\
			MPerS$^*$ (Ours) & DINOv3-LVD & \textbf{82.13}/\textbf{90.19} & \textbf{90.56}/\textbf{95.05} & \textbf{64.92}/\textbf{78.73} & \textbf{79.48}/\textbf{88.57} & \textbf{75.01}/\textbf{85.72} & \textbf{78.42} & \textbf{87.65} \\
			
			\bottomrule
		\end{tabular}%
	}
	\label{tab:Vaihigen}%
\end{table*}%

\begin{figure*}[!t]  % 使用 dblfloatfix
	\centering
	\includegraphics[width=0.8\textwidth]{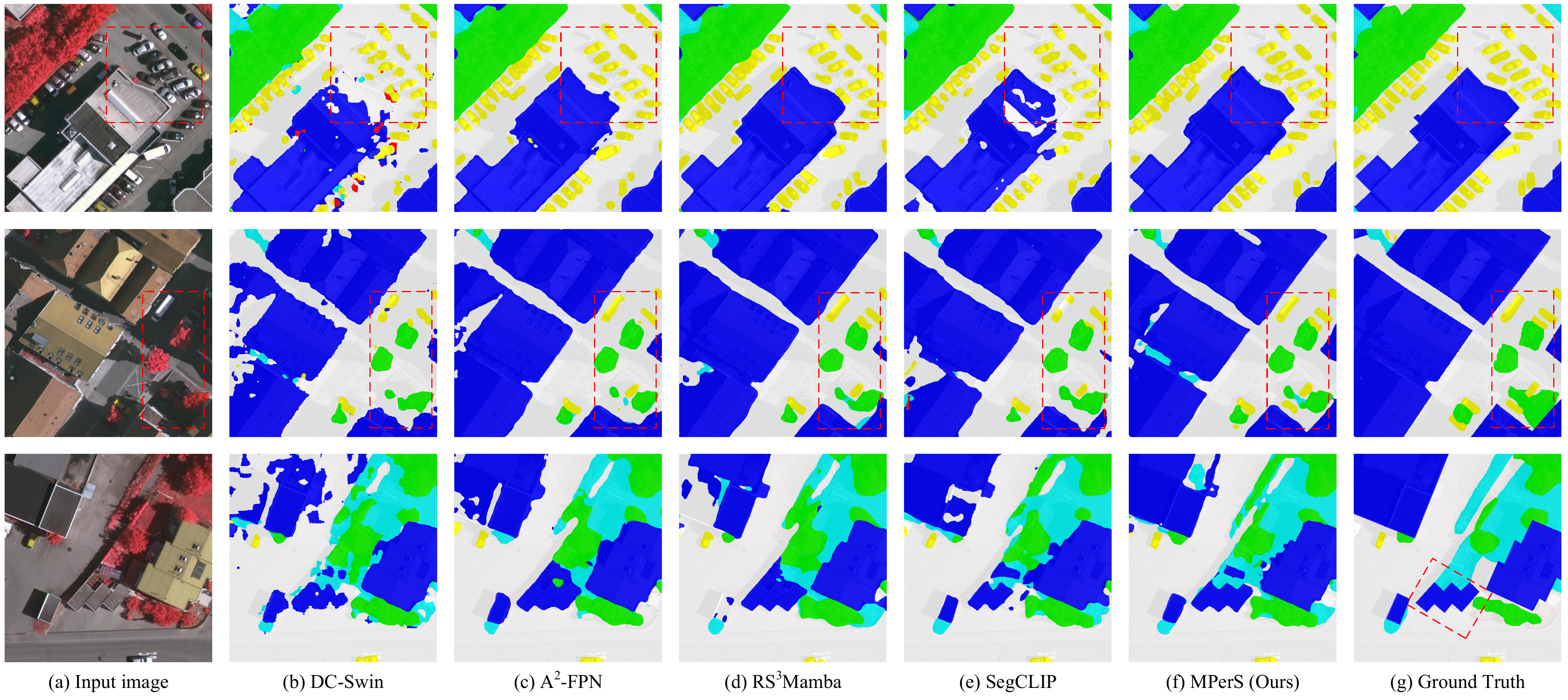}
	\caption{Qualitative visual comparison with state-of-the-art methods on the Vaihingen dataset. Dashed bounding boxes indicate regions where our model produces more precise segmentation results.}
	\label{fig:Seg}
\end{figure*}

\begin{table*}[htbp]
	\centering
	\caption{Quantitative comparison with state-of-the-art methods on the Potsdam dataset. FILM employs a self-designed backbone.}
	\resizebox{\linewidth}{!}{
		\begin{tabular}{c|c|ccccc|cc}
			\toprule
			\multirow{2}[4]{*}{Model} & \multirow{2}[4]{*}{Backbone} & \multicolumn{5}{c|}{Per-class IoU(\%)/F1(\%)} & \multirow{2}[4]{*}{mIoU(\%)} & \multirow{2}[4]{*}{mF1(\%)} \\
			\cmidrule{3-7}          &       & Impervious surface & Building & Low vegetation & Tree & Car  &       &  \\
			\midrule
			MAResUnet & ResNet-34 & 80.70/89.32 & 90.63/95.08 & 67.27/80.43 & 67.21/80.39 & 81.29/89.68 & 77.16 & 86.79 \\
			Unetformer & ResNet-18 & 79.79/88.76 & 88.72/94.02 & 67.21/80.39 & 66.12/79.60 & 81.43/89.76 & 76.65 & 76.65 \\
			DC-Swin & Swin-B & 76.42/86.63 & 85.89/92.41 & 62.19/76.69 & 62.19/76.69 & 76.41/86.63 & 72.24 & 83.51 \\
			A$^2$-FPN & ResNet-18 & 81.60/89.87 & 91.45/95.53 & 69.57/82.06 & 69.36/81.91 & 82.28/90.28 & 78.85 & 87.93 \\
			RS$^3$Mamba & VMamba-T & 81.84/90.02 & 91.26/95.43 & 70.02/82.37 & 69.55/82.04 & 81.94/90.07 & 78.92 & 87.98 \\
			FILM  & -   & 71.99/83.72 & 78.74/88.11 & 61.56/76.20 & 53.43/69.65 & 61.70/76.31 & 65.48 & 65.48 \\
			MetaSegNet & Swin-B & 77.52/87.34 & 88.83/94.08 & 65.17/78.92 & 67.90/80.88 & 73.98/85.04 & 74.68 & 85.25 \\
			SegCLIP & Swin-B & 80.13/88.97 & 89.97/94.72 & 67.81/80.82 & 69.41/81.95 & 81.52/89.82 & 77.77 & 87.25 \\
			\midrule
			MPerS (Ours) & DINOv3-SAT & \underline{82.51}/\underline{90.42} & \underline{91.67}/\underline{95.66} & \underline{70.56}/\underline{82.74} & \underline{70.49}/\underline{82.69} & \underline{82.62}/\underline{90.48} & \underline{79.57} & \underline{88.40} \\
			MPerS (Ours) & DINOv3-LVD & \textbf{83.68}/\textbf{91.12} & \textbf{93.08}/\textbf{96.41} & \textbf{72.36}/\textbf{83.97} & \textbf{72.22}/\textbf{83.87} & \textbf{83.86}/\textbf{91.22} & \textbf{81.04} & \textbf{89.32} \\		
			\bottomrule
		\end{tabular}%
	}
	\label{tab:Potsdam}%
\end{table*}%

\begin{table*}[htbp]
	\centering
	\caption{Quantitative comparison with state-of-the-art methods on the SynDrone dataset. FILM employs a self-designed backbone.}
	\resizebox{\linewidth}{!}{
		\begin{tabular}{c|c|ccccccc|cc}
			\toprule
			\multirow{2}[4]{*}{Model} & \multirow{2}[4]{*}{Backbone} & \multicolumn{7}{c|}{Per-class IoU(\%)/F1(\%)}         & \multirow{2}[4]{*}{mIoU(\%)} & \multirow{2}[4]{*}{mF1(\%)} \\
			\cmidrule{3-9}          &       & Road  & Nature & Person & Vehicle & Construction & Obstacle & Water &       &  \\
			\midrule
			MAResUnet & ResNet-34 & 80.29/89.07 & 88.67/93.99 & 1.34/4.68 & 39.98/57.12 & 57.26/72.82 & 9.63/17.57 & 77.87/87.56 & 50.72 & 60.40 \\
			Unetformer & ResNet-18 & 77.10/87.07 & 82.50/90.41 & 0.40/0.80 & 22.03/36.11 & 45.28/62.33 & 9.66/17.61 & 73.47/84.71 & 44.35 & 54.15 \\
			DC-Swin & Swin-B & 80.24/89.04 & 88.29/93.78 & 0.16/0.31 & 27.89/43.62 & 54.77/70.78 & 9.72/17.71 & 74.17/85.17 & 47.89 & 57.2 \\
			A$^2$-FPN & ResNet-18 & 82.51/90.42 & 89.83/94.64 & 10.25/18.60  & 59.98/74.99 & 61.67/76.29 & 23.00/37.40 & 81.33/89.70 & 58.37 & 68.86 \\
			RS$^3$Mamba & VMamba-T & 82.91/90.66 & 89.29/94.34 & 2.83/5.49 & 53.21/69.46 & 53.21/69.46 & 16.29/28.02 & 81.26/89.66 & 55.17 & 64.71 \\
			FILM  & -     & 76.20/86.49 & 85.13/91.97 & 2.35/5.12 & 3.72/7.17 & 42.24/59.39 & 0.32/0.63 & 80.84/89.41 & 41.54 & 48.60 \\
			MetaSegNet & Swin-B & 84.36/91.52 & \underline{91.90}/\underline{95.78} & 1.54/4.82 &  45.26/62.31 & 72.51/84.07 &  4.62/8.83  & 73.36/84.63 & 53.36 & 61.71 \\
			SegCLIP & Swin-B & \underline{89.82}/\underline{94.64} & 89.82/94.64 & 12.56/22.31 & 61.85/76.43 & \underline{74.90}/\underline{85.65} & 19.97/33.29 & 83.84/91.21 & 62.38 & 71.47 \\
			\midrule
			MPerS (Ours) & DINOv3-SAT & 84.15/92.56 & 91.34/95.47 & \underline{26.96}/\underline{42.47} & \underline{65.08}/\underline{78.85} & 67.00/80.24 & \underline{27.52}/\underline{43.16} & \underline{84.01}/\underline{91.31} & \underline{64.01} & \underline{74.87} \\
			MPerS (Ours) & DINOv3-LVD & \textbf{91.54}/\textbf{95.58} & \textbf{95.72}/\textbf{97.81} & \textbf{34.43}/\textbf{51.23} & \textbf{78.27}/\textbf{87.81} & \textbf{85.52}/\textbf{92.19} & \textbf{35.47}/\textbf{52.36} & \textbf{84.25}/\textbf{91.45} & \textbf{72.17} & \textbf{81.21} \\
			
			\bottomrule
		\end{tabular}%
	}
	\label{tab:SynDrone}%
\end{table*}%

\begin{table}[htbp]
	\centering
	\caption{Ablation of components on the Vaihingen dataset.}
	\resizebox{\linewidth}{!}{
		\begin{tabular}{cccc|ccc}
			\toprule
			Baseline  & LDPE  & LQGA  & DMTE  & OA(\%)   & mIoU(\%)  & F1(\%) \\
			\midrule
			$\checkmark$     & $-$      & $-$      &    $-$   & 87.57 & 73.76 & 84.42 \\
			$\checkmark$     & $\checkmark$     &   $-$    &   $-$    & 87.95 & 74.68 & 85.12 \\
			$\checkmark$     & $\checkmark$     & $\checkmark$     &   $-$    & 88.27 & 76.36 & 86.30 \\
			\midrule
			\rowcolor{gray!20}
			$\checkmark$     & $\checkmark$     & $\checkmark$     & $\checkmark$     & \textbf{88.52} & \textbf{77.01} &\textbf{ 86.74} \\
			\bottomrule
		\end{tabular}%
	}
	\label{tab:components}%
\end{table}%

\subsection{Dataset detail}

\noindent \textbf{Vaihigen Dataset.}
The Vaihingen dataset \cite{waqas2019isaid} consists of 33 true orthophoto (TOP) images captured by advanced airborne sensors, covering an area of approximately 1.38 $\text{km}^\text{2}$ in Vaihingen. The ground sampling distance (GSD) for these images is approximately 9 cm. Each TOP image is composed of three channels: infrared, red, green and Blue (IRGB). Our work uses 12 images for training and 4 for testing. Cropping them into 512$\times$512 patches yields 780 training tiles and 266 testing tiles.

\noindent\textbf{Potsdam Dataset.}
The Potsdam dataset \cite{waqas2019isaid} consists of 38 patches of the same size (6000$\times$6000), all extracted from the TOP mosaic with a GSD of 5 cm. This dataset covers an area of 3.42 $\text{km}^\text{2}$  in Potsdam, featuring dense settlement structures. Our work uses 18 images for training and 6 for testing. Cropping them into 512$\times$512 patches yields 9,522 training tiles and 3,172 testing tiles.

\noindent \textbf{SynDrone Dataset.}
The SynDrone dataset \cite{rizzoli2023syndrone} supports object-level and pixel-wise understanding of urban drone imagery. Data were collected from drone flights over eight towns, with only 90\textdegree ~viewing angle images selected. Each sample is a 1920×1080 three-channel RGB image. SynDrone provides fine-grained annotations. In our work, we use the coarse label set of seven common categories as in Table~\ref{tab:SynDrone}.
Cropping all chosen images into 512$\times$512 patches yields 10,080 training tiles and 3,360 testing tiles.

\subsection{Experimental Settings}

\subsubsection{Implementation detail}
We develop our model using the PyTorch framework. For visual feature extraction, the backbone employs a frozen DINOv3 to perform encoding. During the experiments on the Vaihingen dataset, we adopt the DINOv3 backbone pretrained on the SAT and LVD datasets, including both the 0.3B distilled weights and the full 7B weights. Meanwhile, for Potsdam and SynDrone, only the faster-training SAT-0.3B and LVD-0.3B distilled weights are used. For text feature extraction, we use the text encoder component of CLIP, which remains frozen during training. The initial learning rate is set to 0.001. A multi-step learning rate scheduler is employed for optimization. We train the model using the AdamW optimizer with a batch size of 8. We use LLaVA-v1.5-7B, ChatGPT-4o, and Qwen2.5-3B to generate captions for RS-Scene images, respectively. The experiments are conducted using an NVIDIA A800 80 GB GPU to train our model.

\subsubsection{Evaluation Metrics}
We compute the Intersection-over-Union (IoU) and F1 score for all classes to thoroughly assess the segmentation accuracy of the model for each type of RS land cover. We also report the mean IoU (mIoU) and mean F1 score (mF1) for all classes to provide an intuitive comparison of different methods on RS-Scene segmentation performance. In subsequent ablation experiments, we further use Overall Accuracy (OA) to evaluate the effects of different components.

\subsection{Comparison with state-of-the-art}
To avoid the limitations of a single comparison, we compare our method with several state-of-the-art single-modal and text-attention multimodal segmentation approaches, including MAResUNet \cite{li2021multistage}, UNetFormer \cite{wang2022unetformer}, DC-Swin \cite{wang2022novel}, A$^2$-FPN \cite{hu2021a2}, RS$^3$Mamba \cite{ma2024rs}, MetaSegNet \cite{wang2024metasegnet}, FiLM\cite{zhao2024image}, and SegCLIP\cite{zhang2024segclip}. Among them, MetaSegNet, FiLM , and SegCLIP  are the latest multimodal text-guided semantic segmentation methods, while the others are high-performing segmentation approaches on RS datasets. For more qualitative analyses, please refer to Section 3 of the supplementary material.

\noindent \textbf{Results on the Vaihigen Dataset.} The experimental results of various methods on the Vaihingen dataset in Table~\ref{tab:Vaihigen}. It can be observed that, compared with other semantic segmentation methods, MPerS significantly improves the mF1 and mIoU metrics. Compared with using the 0.3B distilled DINOv3 weights from LVD and SAT, MPerS achieves consistent performance improvements when adopting the full 7B weights of DINOv3 trained on LVD and SAT. Our method achieves the best mF1 and mIoU of 78.42\% and 87.65\%, respectively, significantly outperforming other approaches. To better facilitate practical applications, we compare other methods against MPerS using the LVD-0.3B weights, where the segmentation accuracy reaches an mIoU of 77.10\% and an F1 score of 86.79\%.

Compared with the second-best single-modality method RS$^3$Mamba, MPerS improves mIoU and mF1 by 3.02\% and 2.08\%, respectively. Against FiLM, which employs a direct text-image fusion, our method shows notable gains on the ``car'' category. Compared with the latest SegCLIP, mIoU and mF1 increase by 4.87\% and 5.52\%, respectively. Relative to MetaSegNet, which uses captions from a single LLM, improvements of 10.72\% and 5.52\% in mIoU and mF1 are observed. Notably, MPerS maintains high accuracy for large objects while significantly improving ``Tree'' and ``Low vegetation''. The DMTE module effectively guides the model to small objects, yielding a 6.75\% IoU gain for dense small objects like cars compared to the second-best method.

\noindent \textbf{Results on the Potsdam Dataset.} The Potsdam and Vaihingen datasets have similar object categories, but Potsdam exhibits more complex terrain and denser overlapping objects. 
As shown in Table~\ref{tab:Potsdam}, compared with other methods, MPerS still significantly outperforms other models in segmentation performance using the LVD-0.3B distilled weights. 
Compared with the second-best RS$^3$Mamba, mF1 and mIoU improve by 2.12\% and 1.34\%, respectively, and compared with SegCLIP, they increase by 5.52\% and 4.87\%. High-quality text guidance further enhances the model’s attention to ``Low vegetation'' and ``Tree'', yielding IoU gains of 2.79\% and 2.67\% over the next-best methods.

\noindent \textbf{Results on the SynDrone Dataset.} Compared with the Potsdam dataset, SynDrone includes both urban and rural scenes, featuring more complex categories and more challenging foreground segmentation problems. 
On SynDrone, SAT-0.3B shows limited improvement, while LVD-0.3B still achieves strong segmentation performance. 
As shown in Table~\ref{tab:SynDrone}, our method still achieves excellent segmentation performance for foreground objects with few samples and extremely small sizes, such as ``Person" and ``Obstacle". 
MPerS achieves IoU scores of 34.43\% and 35.47\% for the two RS-Scene categories. Compared with SegCLIP, the IoU for the ``Vehicle’’ class rises from 61.85\% to 78.27\%. In addition, it maintains strong performance on large-area objects, achieving IoUs of 91.54\% and 95.72\% for the ``road’’ and ``natural’’ classes, respectively.

\noindent \textbf{Segmentation Results Visualization.} As illustrated in the segmentation results in Fig.~\ref{fig:Seg}, our method achieves more complete and coherent segmentation of large land-cover objects, while significantly improving the segmentation accuracy of dense, small-scale objects such as ``Cars''. It produces finer boundaries and fewer segmentation errors. Additional visualizations on other datasets are provided in Supplementary Section 4.

\subsection{Ablation study}

\subsubsection{Main components ablation}

\noindent\textbf{Component I: DINOv3 Baseline.} Our model employs a frozen DINOv3 as the vision encoder. As shown in Table~\ref{tab:components}, even without external components, DINOv3 can effectively extract visual representations in dense RS-Scenes, achieving performance comparable to traditional methods. As shown in Table~\ref{tab:DINOv3 weights}, we further analyze the differences between the SAT and LVD weights. The DINOv3 weights trained on the satellite dataset perform notably worse than those trained on the web-based dataset in RS task. Therefore, for practical considerations, we primarily employ the LVD-0.3B distilled weights for model construction and ablation analysis.

\begin{table}[htbp]
	\centering
	\caption{Quantitative analysis with DINOv3 weights.}
	\resizebox{\linewidth}{!}{
		\begin{tabular}{l|ccc|c}
			\toprule
			\multicolumn{1}{c|}{Method} & Vaihingen & Potsdam & SynDrone & Average \\
			\midrule
			Baseline SAT-0.3B ~~\raisebox{-0.5ex}{\includegraphics[width=0.4cm]{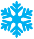}} & 69.48 & 73.85 & 51.68 & 65.01 \\
			Baseline LVD-0.3B ~\raisebox{-0.6ex}{\includegraphics[width=0.4cm]{pic/snowflake.pdf}}& \textbf{73.76} & \textbf{77.11} & \textbf{56.24} & \textbf{69.04} \\
			\bottomrule
		\end{tabular}%
	}
	\label{tab:DINOv3 weights}%
\end{table}%

\noindent\textbf{Component II: Lite Detail Prior Encoder.} Although DINOv3 performs well in recognizing land-covers in dense RS-Scenes, the extracted features are primarily general visual representations. Due to the lack of domain priors for specific RS scenes, we design the LDPE to extract RS-specific detail prior features. As shown in Table~\ref{tab:components}, the results indicate that the LDPE contributes positively to the performance. The LDPE is composed of lightweight convolutions module and partially functions as a DINOv3 adapter, enabling it to better segment RS-Scence.

\noindent\textbf{Component III: Linguistic Query Guided Attention.} Ablation experiments on QGCA are conducted using high-quality caption generated by MLLM. As shown in Table~\ref{tab:components}, incorporating textual information notably enhances segmentation performance, improving OA, mIoU, and mF1 by 0.7\%, 2.6\%, and 1.88\% over the baseline. This aims to demonstrate the effectiveness of QGCA in text-guided feature extraction and fusion.

\noindent\textbf{Component IV: Dynamic MixExperts Text Encoder.} As shown in Table~\ref{tab:components}, the DMTE effectively leverages the textual semantic information of RS-Scenes perceived by different MLLM experts to guide visual features. In our study, a mixture-of-experts (MOE) structure constructed from only three types of MLLMs achieves robust performance. The architecture and MoE design warrant further investigation.

\begin{table}[t!]
	\centering
	\caption{Effectiveness analysis of remote sensing scene caption.}
	\resizebox{\linewidth}{!}{
		\begin{tabular}{llcc}
			\toprule
			\multicolumn{2}{l}{Method} & mIoU(\%)  & mF1(\%) \\
			\midrule
			\multicolumn{2}{l}{RS$^3$Mamba} & 74.08 & 84.71 \\
			\multicolumn{2}{l}{baseline} & 73.76 & 84.42 \\
			\multicolumn{2}{l}{MPerS + simple inquiry prompt caption} & 75.59 & 85.81 \\
			\multicolumn{2}{l}{MPerS + multi-prompts single MLLM caption} & 76.36 & 86.30 \\
			\multicolumn{2}{l}{MPerS + DMTE(single MLLM caption)} & \underline{76.41} & \underline{86.33} \\
			\midrule
			\rowcolor{gray!20}
			\multicolumn{2}{l}{MPerS (Ours)} & \textbf{77.10}  & \textbf{86.79} \\
			\bottomrule
		\end{tabular}%
	}
	\label{tab:captions}%
\end{table}%

\subsubsection{Caption Effectiveness Analysis}
This study further investigates how the quality of captions affects the model's performance. Table~\ref{tab:captions} shows that, for MLLMs, captions generated from RS-specific prompts are more effective than those from simple prompts. As shown in Tables~\ref{tab:components} and~\ref{tab:captions}, the multi-expert design is more effective in extracting meaningful textual semantic information for downstream tasks. The effects of different text encoders and the visualization analyses of captions are presented in Sections 1 and 2 of the supplementary material, respectively.

\begin{table}[htbp]
	\centering
	\caption{Quantitative Efficiency Comparison.}
	\resizebox{0.7\linewidth}{!}{
		\begin{tabular}{c|c|c}
			\toprule
			Method & Backbone & Inference time(ms) $\downarrow$\\
			\midrule
			MAResUnet & ResNet-34 & 42.99 \\
			Unetformer & ResNet-18 & 147.7 \\
			DC-Swin & Swin-B & 283.39 \\
			A2FPN & ResNet-18 & 21.85 \\
			RS$^3$Mamba & VMamba-T & \underline{25.8} \\
			MetaSegNet & Swin-B & \textbf{14.01} \\
			SegCLIP & Swin-B & 29.32 \\
			\midrule
			MPerS (Ours) & DINOv3 & 55.52 \\
			\bottomrule
		\end{tabular}%
	}
	\label{tab:Efficiency}%
\end{table}%

\subsubsection{Complexity Analysis}
Table~\ref{tab:Efficiency} presents the inference time of the segmentation models evaluated in this study on a single 512$\times$512 RS-Scene. Although the inclusion of DINOv3 and CLIP inevitably reduces the inference speed of our method, the impact remains acceptable for RS tasks. MPerS method is not only feasible for RS semantic segmentation tasks but also holds potential for further exploration and applications.
\section{Conclusion}

In this study, we have proposed MPerS to tackle the challenge of effectively utilizing captions generated by multimodal large language models (MLLMs) with precise perceptual understanding to guide multimodal dense remote sensing scene (RS-Scene) segmentation.
It primarily leverages DINOv3 to extract vision features and the CLIP text encoder to extract textual semantic information. 
To ensure that MLLMs positively perceive RS-Scene, we design multi-perspective prompts and check strategy. 
We have explored how to extract and organize the effective textual information perceived by different MLLM experts for downstream segmentation tasks.
Notably, each component of MPerS (CLIP, DINOv3, and the MixExperts) is modular and can be replaced or iteratively updated. Experiments show that our model achieves state-of-the-art performance.
We hope that our study provides a new solution for exploring how multimodal large models accurately perceive RS scenes and facilitate practical RS interpretation.
 
{
    \small
    \bibliographystyle{ieeenat_fullname}
    \bibliography{main}
}


\end{document}